\documentclass[conference]{IEEEtran}
\usepackage{times}
\usepackage{cite}
\usepackage{amsmath,amssymb,amsfonts}
\usepackage{algorithmic}
\usepackage{graphicx}
\usepackage{textcomp}
\usepackage{setspace}
 \usepackage{booktabs,array,colortbl,graphicx,diagbox, multirow}  
\usepackage[table]{xcolor}

\usepackage{epsfig}
\usepackage{times} 
\usepackage{svg}
\usepackage[linesnumbered,ruled,vlined]{algorithm2e} 

\usepackage[nolist, nohyperlinks]{acronym}

\usepackage{bm}
\usepackage{breqn}
\usepackage{amsmath}
\usepackage{amssymb}
\usepackage{tabularx}

\usepackage{amsthm}

\newcommand{\Z}{\mathbb{Z}}

\newcommand{\R}{\mathbb{R}}

\newcommand{\xbf}{\mathbf{x}}
\newcommand{\ybf}{\mathbf{y}}

\newcommand{\Ebb}{\mathbb{E}}

\newcommand{\Bcal}{\mathcal{B}}
\newcommand{\Ccal}{\mathcal{C}}
\newcommand{\Dcal}{\mathcal{D}}

\newcommand{\Fcal}{\mathcal{F}}

\newcommand{\Ical}{\mathcal{I}}

\newcommand{\Kcal}{\mathcal{K}}
\newcommand{\Lcal}{\mathcal{L}}

\newcommand{\Ncal}{\mathcal{N}}

\newcommand{\Pcal}{\mathcal{P}}

\newcommand{\Rcal}{\mathcal{R}}
\newcommand{\Scal}{\mathcal{S}}
\newcommand{\Tcal}{\mathcal{T}}

\newcommand{\Xcal}{\mathcal{X}}

\newcommand{\eqn}[1]{\begin{align} #1 \end{align}}
\newcommand{\eqnN}[1]{\begin{align*} #1 \end{align*}}
\newcommand{\seqn}[2][]{
\begin{subequations}
    #1
\begin{align} #2 \end{align}
\end{subequations}
}

\newcommand{\argmin}[1]{\underset{\substack{#1}}{\text{argmin}}}
\newcommand{\argmax}[1]{\underset{\substack{#1}}{\text{argmax}}}

\acrodef{DDPM}[DDPM]{Denoising Diffusion Probabilistic Model}
\acrodef{D3PM}[D3PM]{Discrete Denoising Diffusion Probabilistic Model}
\acrodef{MBD}[MBD]{Model-Based Diffusion}
\acrodef{MBD2}[MBD2]{Model-Based Discrete Diffusion}
\acrodef{DMBD}[Discrete MBD]{Discrete Model-Based Diffusion}

\theoremstyle{plain}
\newtheorem{theorem}{Theorem}

\newtheorem{lemma}{Lemma}

\theoremstyle{definition}
\newtheorem{assumption}{Assumption}
\newtheorem{remark}{Remark}

\makeatletter
\let\NAT@parse\undefined
\makeatother

\usepackage[bookmarks=true]{hyperref}
\usepackage{cleveref}

\crefformat{problem}{Problem~#2#1#3}
\crefformat{assumption}{Assumption~#2#1#3}
\crefformat{prop}{Proposition~#2#1#3}
\usepackage[table]{xcolor}

\usepackage[numbers]{natbib}
\usepackage{multicol}

\begin{document}

\title{Discrete Model-Based Diffusion for Combinatorial Optimization in Multi-Robot Task Allocation}

% You will get a Paper-ID when submitting a pdf file to the conference system
\author{\authorblockN{Haejoon Lee\textsuperscript{1},
Thomas Marshall Vielmetti\textsuperscript{1},
Xinyi Wang\textsuperscript{2}, and
Dimitra Panagou\textsuperscript{1}}
\authorblockA{\textsuperscript{1}Robotics Department, University of Michigan,
Ann Arbor, MI, USA\\
Email: {\tt \{haejoonl, mvielmet, dpanagou\}@umich.edu}}
\authorblockA{\textsuperscript{2}Department of Mechanical Engineering, Oakland University,
Rochester, MI, USA\\
Email: {\tt xinyiwang@oakland.edu}}}

%\author{\authorblockN{Michael Shell}
%\authorblockA{School of Electrical and\\Computer Engineering\\
%Georgia Institute of Technology\\
%Atlanta, Georgia 30332--0250\\
%Email: mshell@ece.gatech.edu}
%\and
%\authorblockA{Twentieth Century Fox\\
%\and
%\authorblockN{James Kirk\\ and Montgomery Scott}
%\authorblockA{Starfleet Academy\\
%San Francisco, California 96678-2391\\

%Homer Simpson\authorrefmark{2},
%Montgomery Scott\authorrefmark{3} and
%Eldon Tyrell\authorrefmark{4}}
%Telephone: (800) 555--1212, Fax: (888) 555--1212}
%\authorblockA{\authorrefmark{4}Tyrell Inc., 123 Replicant Street, Los Angeles, California 90210--4321}}

\maketitle

\begin{abstract} 
Decision making in multi-robot systems often involves discrete decision variables, leading to NP-hard combinatorial optimization problems that must be solved within seconds for near real-time implementation. Denoising diffusion has emerged as a powerful tool for such problems, but existing approaches face a trade-off: Model-Based Diffusion (MBD) incorporates diffusion in sampling-based optimization (SBO) without training but only over continuous spaces, while existing diffusion-based combinatorial solvers operate over discrete spaces but require offline training. To this end, we propose Discrete Model-Based Diffusion (Discrete MBD), a training-free SBO method that directly optimizes discrete variables by approximating categorical reverse transition distributions through Monte Carlo sampling. We establish a finite-step lower bound on the probability of finding a global minimizer and show that this bound converges to one as the number of denoising steps increases. Extensive experiments on generic integer programs and heterogeneous multi-robot coalition formation and scheduling demonstrate that Discrete MBD outperforms SBO baselines in nearly all settings while achieving solution quality competitive with that of mixed-integer programming and reinforcement learning methods at a fraction of their computational cost. [Project Page\footnote{\href{https://joonlee16.github.io/discrete-mbd.github.io/}{https://joonlee16.github.io/discrete-mbd.github.io/}}] [Web Demo\footnote{\href{https://joonlee16.github.io/playground/discrete-mbd}{https://joonlee16.github.io/playground/discrete-mbd}}]

\end{abstract}
\IEEEpeerreviewmaketitle

\section{Introduction}
\label{sec:intro}

Many decision-making problems in robotic systems involve inherently discrete choices, such as assigning tasks to robots~\cite{schillinger2018simultaneous, fu2023robust, weintraub2025virtual}, selecting modes of hybrid systems~\cite{marcucci2020warm, hogan2020reaective}, and avoiding obstacles~\cite{ioan2021mixed, kurtz2023temporal}. These decisions give rise to combinatorial optimization problems that are often NP-hard, with solution spaces growing exponentially with problem size, making high-quality solutions difficult to obtain within the tight time budgets of online operation.

\begin{figure}
    \centering
    \includegraphics[width=\linewidth]{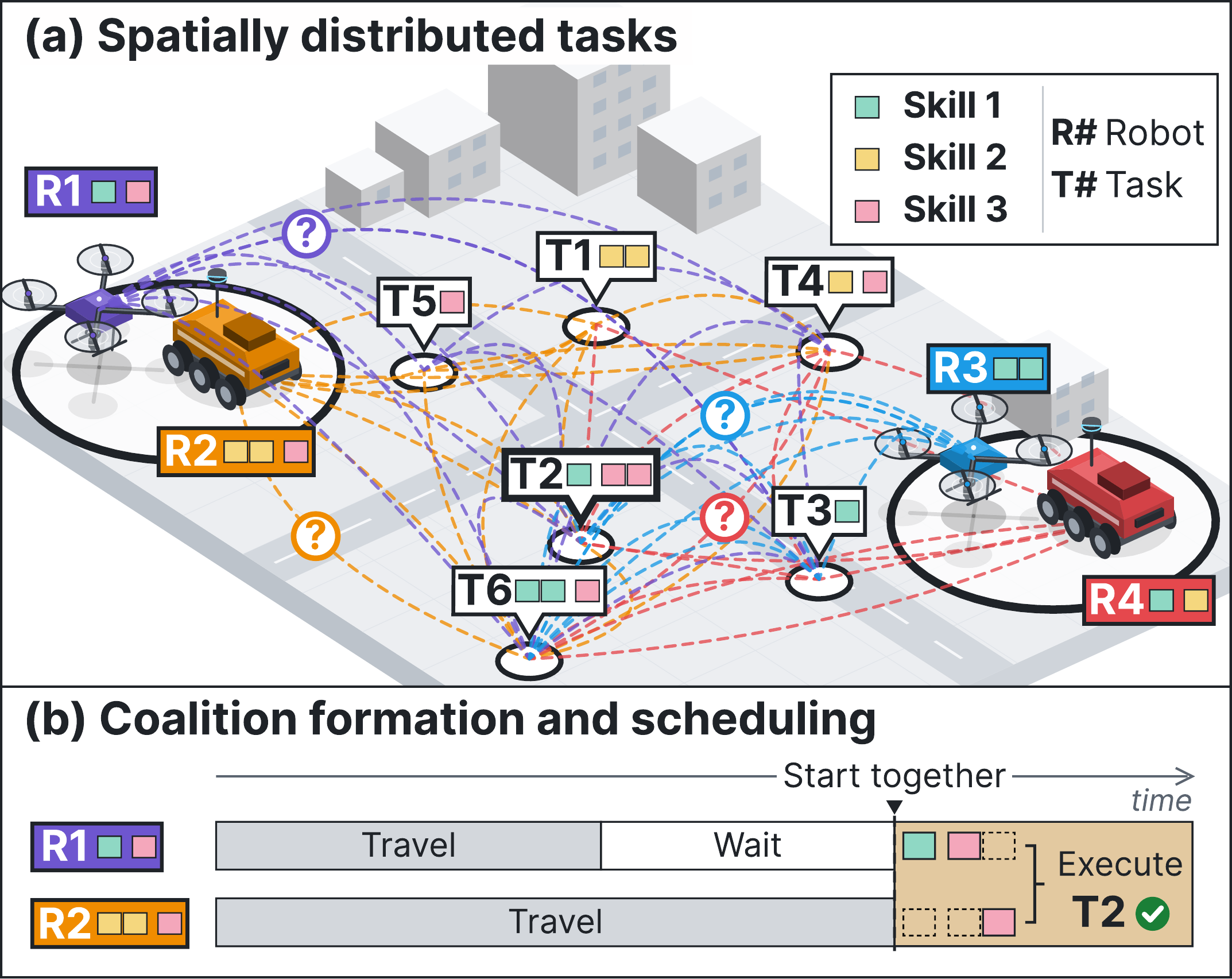}
    \caption{Heterogeneous MRTA problem, which jointly addresses coalition formation and scheduling. Each robot has multiple skills at different capability levels (one block per level), and each task requires a minimum combined capability level for each specified skill. Robots must form coalitions and schedule routes to minimize the makespan. (a)~Dashed curves show candidate routes. (b)~Example coalition: R1 and R2 jointly meet T2’s requirement, so R1 waits for R2 to arrive before the task starts.}
    \label{fig:hero}
\end{figure}

A representative example, used throughout this paper, is the heterogeneous multi-robot task allocation (MRTA) problem, where robots with different skills, each at a different capability level, must complete spatially distributed tasks as quickly as possible~\cite{fu2023robust,aswale2023heterogeneous,dai2025heterogeneous} (\Cref{fig:hero}). Specifically, we consider joint coalition formation and scheduling, where each task requires \emph{tight coordination}~\cite{rizk2019cooperative}: task execution begins only once a coalition with sufficient combined skill capabilities gathers at the task location. Robots must therefore jointly decide which robots serve each task (\emph{coalition formation}) and when each robot executes its tasks (\emph{scheduling}) to minimize makespan. This places the problem in the \emph{single-task robots, multi-robot tasks, time-extended assignment, cross-schedule dependencies} (XD~[ST-MR-TA]) category~\cite{korsah2013comprehensive}, with a combinatorial search space that grows rapidly with the numbers of robots and tasks, and complexity of the skill requirements. Since new tasks, robot dropouts, or travel delays may arise unexpectedly~\cite{dai2025heterogeneous}, robots must replan within seconds to dynamically adapt to such changing conditions.

There exist numerous approaches to these combinatorial optimization problems. Exact methods formulate problems as optimization programs~\cite{fu2023robust,chen2025new}  and employ techniques such as Branch-and-Bound and cutting-plane methods~\cite{nemhauser1988integer}, but scale poorly and become computationally expensive for large problems. Metaheuristics, such as simulated annealing (SA)~\cite{bertsimas1993simulated}, genetic algorithms~\cite{bean1994genetic,mirjalili2018genetic}, ant colony optimization~\cite{babincsak2023ant}, and large neighborhood search~\cite{pisinger2018large,wilde2023designing}, offer faster solutions but require problem-specific local search operators or heuristics. Learning-based methods~\cite{wang2020heterogenous,sun2023difusco,dai2025heterogeneous, khosravi2026diffusion} enable fast inference but require offline training and may generalize poorly to problem structures unconsidered during the training.

Sampling-based optimization (SBO) offers an alternative. SBO methods iteratively sample candidate solutions from a probability distribution and refine the distribution to concentrate its search on promising regions, without problem-specific local search algorithms or offline training. While closely related methods are known as \textit{model-based search} algorithms in combinatorial optimization~\cite{zlochin2004model}, we follow the robotics and control literature~\cite{williams2018information,yi2026global} and adopt the term SBO in this paper. Classical SBO methods for discrete problems include the cross-entropy method (CEM)~\cite{rubinstein1999cross} and sequential Monte Carlo (SMC)~\cite{del2006sequential}. However, these methods struggle to explore large solution spaces with multimodal objective landscapes, i.e., objective functions with many local optima: CEM is prone to premature mode collapse, while SMC suffers from particle degeneracy that reduces sample diversity.

Meanwhile, \acf{MBD}~\cite{pan2024model} has emerged as a promising SBO method that adapts the diffusion process of \ac{DDPM}~\cite{ho2020denoising} to optimization. \ac{MBD} uses the known objective function to estimate the score function through Monte Carlo sampling, eliminating the need for offline neural network training. By gradually concentrating samples on low-cost regions through the diffusion process, \ac{MBD} outperforms other SBO methods on nonconvex, non-differentiable problems~\cite{pan2024model, lee2026distributed}.

Diffusion-based search is also appealing for combinatorial optimization, where solution spaces are often large and highly multimodal~\cite{sun2023difusco, khosravi2026diffusion}. However,~\ac{MBD} denoises in a continuous space and thus cannot directly produce discrete decisions. Its solutions must be rounded, which degrades solution quality~\cite{sun2023difusco} and may lead to constraint violations. On the other hand, existing diffusion-based combinatorial optimization solvers~\cite{sun2023difusco,li2024fast,sanokowski2025scalable,lei2026boosting, khosravi2026diffusion} denoise directly over discrete variables using discrete diffusion models~\cite{hoogeboom2021argmax,austin2021structured}. However, they rely on a neural network to learn the denoising process offline.

To bridge this gap, we propose~\acf{DMBD}, a training-free SBO method that performs denoising diffusion directly over discrete decision variables, modeling each as a categorical random variable. At each denoising step, \ac{DMBD} approximates the categorical reverse transition distribution with samples weighted by the known problem formulation, progressively steering the search toward high-quality solutions. The contributions of this paper are as follows:
\begin{itemize}
\item We propose Discrete MBD, an SBO method for combinatorial optimization that denoises directly over discrete decision variables by approximating the reverse transitions of discrete diffusion with Monte Carlo sampling.
\item We establish a finite-step probabilistic convergence guarantee for Discrete MBD (\Cref{thm:convergence}) that characterizes its exploration-exploitation trade-off and shows that the probability of finding a global minimizer approaches one as the number of denoising steps increases.
\item We present a priority-based encoding scheme, inspired by~\cite{bean1994genetic}, that guarantees every sample decodes into a feasible solution for heterogeneous MRTA, improving sampling efficiency.
\item We evaluate our method on numerous generic unconstrained integer programs and heterogeneous MRTA problems.
\end{itemize}

\section{Preliminaries}
We denote the $k$-dimensional column vector of ones by $\mathbf{1}_k$. For matrices $A$ and $B$, let $A\odot B$ denote their elementwise product. For a matrix $A$, we denote by $A_i$ its $i$-th row and by $[A]_{ij}$ the element in row $i$ and column $j$. For a vector $a$, $[a]_i$ denotes its $i$-th element. For vectors $a,b$ of the same size, $a\succeq b$ means $[a]_{i}\ge[b]_{i}$ for all $i$. We denote the sets of positive and nonnegative integers by $\Z_{>0}$ and $\Z_{\ge0}$, respectively. For a set $\Scal$ and $I_1,I_2\in\Z_{>0}$, $\Scal^{I_1}$ denotes the set of $I_1$-tuples with entries in $\Scal$, and $\Scal^{I_1\times I_2}$ the set of $I_1\times I_2$ matrices with entries in $\Scal$.

We consider $N\in\Z_{>0}$ discrete or categorical decision variables indexed by $\Ncal=\{0,\dots,N-1\}$, each taking one of $K\in\Z_{>0}$ categories in $\Kcal=\{0,\dots,K-1\}$. A solution can be represented by a categorical decision vector $\ybf\in\Kcal^N$, where $\ybf_r$ denotes the category of variable $r$. Equivalently, it can be represented by a one-hot matrix $\xbf$ in
\eqn{\Xcal_{N,K}=\left\{\xbf\in\{0,1\}^{N\times K} \mid \xbf\mathbf{1}_K=\mathbf{1}_N\right\},}
where $[\xbf]_{rk}=1$ if and only if $\ybf_r=k$. The two representations are related by $\ybf_r=\sum_{k=0}^{K-1}k\,[\xbf]_{rk}$. Then we can formulate an integer program with a cost function $J:\Xcal_{N,K}\to\R$ and constraint functions $g_d:\Xcal_{N,K}\to\R$, $d=0,\dots,D-1$:
\seqn[\label{eq:original_problem}]{
\Xcal^\star:=\argmin{\xbf\in\Xcal_{N,K}}& J(\xbf) \\
\text{s.t.}& \ g_d(\xbf)\leq 0, \ d = 0,\ldots, D-1,
}
whose feasible set is
\eqn{\Ccal=\left\{\xbf\in\Xcal_{N,K} \mid g_d(\xbf)\leq 0, \ \forall d\in\{0,\ldots,D-1\}\right\}.}

\begin{remark}
    Note that~\eqref{eq:original_problem} covers a broad class of problems, since a category can encode any finite set of options available to a decision variable. For example, suppose each of $N$ robots must choose an order in which to perform $T$ tasks. Each category can then represent one of the $K=T!$ permutations of the tasks.
    Under a fixed enumeration of these sequences, $\ybf_r\in\Kcal$ indicates the specific task sequence performed by robot $r$. With $T=3$, we have $K=6$ permutations: $(0,1,2),(0,2,1),(1,0,2),(1,2,0),(2,0,1),(2,1,0)$, which can be indexed by $\Kcal=\{0,\dots,5\}$, so $\ybf_r=0$ means robot $r$ performs tasks~0, 1, and~2 in that order.
\end{remark}

\subsection{Sampling-Based Optimization}
Solving~\eqref{eq:original_problem} directly quickly becomes intractable as the problem size grows, since the solution space $\Xcal_{N,K}$ contains $K^N$ elements and thus grows exponentially with $N$. Sampling-based optimization (SBO) offers an alternative by recasting~\eqref{eq:original_problem} as probabilistic inference over the target distribution
\eqn{
\label{eq:target_dist}
p_0(\xbf) \propto p_{\Ccal}(\xbf)\,p_J(\xbf)
}
with 
\eqn{
p_{\Ccal}(\xbf) := \mathbb{I}\{\xbf\in\Ccal\},
\quad
p_J(\xbf) := \exp\!\left(-\frac{J(\xbf)}{\lambda}\right),
}
where $p_{\Ccal}$ enforces feasibility and $p_J$ puts more mass on $\xbf$ with lower costs~\cite{rubinstein1999cross, pan2024model}.

Recently, \ac{MBD}~\cite{pan2024model} adapted the denoising process of \ac{DDPM}~\cite{ho2020denoising} to SBO in a \textit{training-free manner} by exploiting the known target distribution~\eqref{eq:target_dist}. Starting from high-variance Gaussian distributions, \ac{MBD} progressively concentrates on low-cost regions through a sequence of smoothed distributions defined by its noise schedule, which makes it effective on nonconvex, non-differentiable problems~\cite{pan2024model, yi2026global}.

\begin{figure}
    \centering
\includegraphics[width=0.9\linewidth]{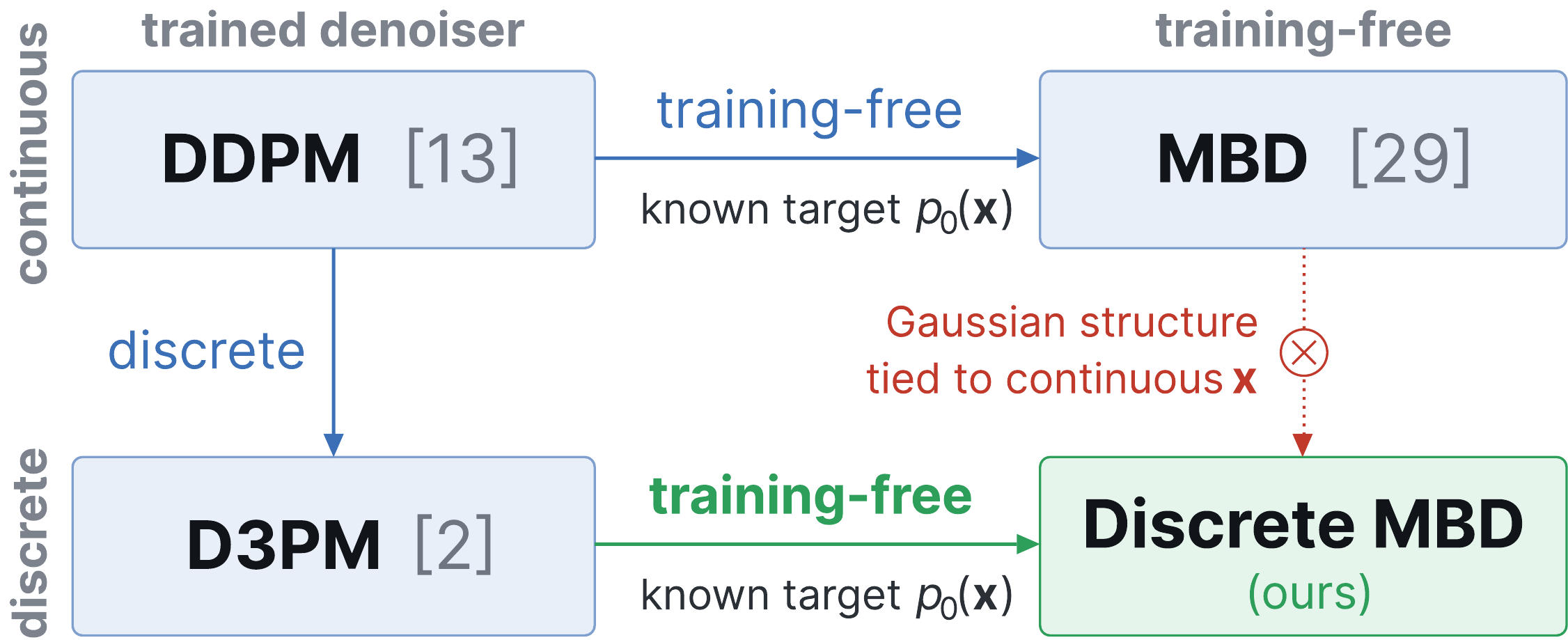}
    \caption{Positioning of \ac{DMBD}. \ac{MBD}~\cite{pan2024model} incorporates the Gaussian diffusion of \ac{DDPM}~\cite{ho2020denoising} into a training-free optimizer with a known target $p_0(\mathbf{x})$. Extending \ac{MBD} directly to discrete
solution spaces is structurally challenging, since its Gaussian diffusion processes
are tied to continuous $\mathbf{x}$ (red arrow). Instead, we start from \ac{D3PM}~\cite{austin2021structured},
the categorical counterpart of \ac{DDPM}, and make it training-free following the same principle as~\ac{MBD}.}
    \label{fig:diagram}
\end{figure}

While the same property is also promising for combinatorial optimization problems like~\eqref{eq:original_problem}, the Gaussian structure of \ac{MBD} is inherently tied to continuous variables and cannot directly accommodate discrete or categorical solution spaces. This structural limitation motivates a different approach: inspired by how~\ac{MBD} incorporates the Gaussian diffusion process of~\ac{DDPM} into training-free optimization with a known target distribution, we similarly adapt the categorical diffusion process of~\ac{D3PM}~\cite{austin2021structured}, the discrete counterpart of~\ac{DDPM}, to training-free optimization by incorporating~\eqref{eq:target_dist} (\Cref{fig:diagram}). The resulting training-free SBO method progressively refines solutions within the discrete solution space.

\subsection{\acf{D3PM}}
\label{subsec:d3pm}
\ac{D3PM}~\cite{austin2021structured} extends the Gaussian diffusion framework of~\ac{DDPM}~\cite{ho2020denoising} to categorical random variables. It defines a discrete-time Markov chain over a finite categorical state space and learns to reverse it.

To describe distributions over $\Xcal_{N,K}$, let $P\in[0,1]^{N\times K}$ be a row-stochastic matrix with rows $P_r$. For $\xbf\in\Xcal_{N,K}$, let $\operatorname{Cat}(\xbf_r;P_r)$ denote the probability mass of its $r$-th row $\xbf_r$ under the categorical distribution with probability row vector $P_r$. We consider distributions over $\mathcal X_{N,K}$ that factorize over rows:
\eqn{\label{eq:cat}
\operatorname{Cat}(\xbf;P) := \prod_{r=0}^{N-1}\operatorname{Cat}(\xbf_r;P_r).
}
\ac{D3PM} consists of a forward and a reverse process, which we describe next.

\textbf{Forward (noising) process:}
The forward process is a fixed Markov chain
\eqn{
q\big(\xbf^{(1)},\ldots,\xbf^{(M)} \mid \xbf^{(0)}\big)
=
\prod_{i=1}^{M} q\big(\xbf^{(i)} \mid \xbf^{(i-1)}\big),
}
that progressively corrupts $\xbf^{(0)}\sim p_0$ into categorical noise over $M$ noising steps. Let $Q_i\in[0,1]^{K\times K}$ be the row-stochastic transition matrix at step $i\in\{1,\dots,M\}$, whose element $[Q_i]_{ab} = q\big([\xbf^{(i)}]_{rb}=1 \mid [\xbf^{(i-1)}]_{ra}=1\big)$ is the probability that the categorical variable $r\in\Ncal$ transitions from category $a$ to category $b$. Then,
\eqn{
\label{eq:one_step_forward}
q\big(\xbf^{(i)} \mid \xbf^{(i-1)}\big) = \operatorname{Cat}\big(\xbf^{(i)};\, \xbf^{(i-1)}Q_i\big),
}
applies the same transition to every row independently. Defining the cumulative transition matrix $\overline{Q}_i:=Q_1Q_2\cdots Q_i$ with $\overline{Q}_0:=I$, the Markov property gives the $i$-step marginal
\eqn{
\label{eq:i_step_forward}
q\big(\xbf^{(i)} \mid \xbf^{(0)}\big)  = \operatorname{Cat}\big(\xbf^{(i)};\, \xbf^{(0)}\overline{Q}_i\big),
}
which allows us to sample $\xbf^{(i)}$ directly from $\xbf^{(0)}$.

In~\cite{austin2021structured}, the rows of $\overline{Q}_i$ are required to converge to a known stationary distribution as $i\to\infty$, for which sufficient conditions are given in~\cite[A.1]{austin2021structured}. In this work, we construct $Q_i$ to satisfy the following assumption:

\begin{assumption}\label{assum:schedule}
The forward transition matrix is uniform,
$Q_i=(1-\beta_i)I+\frac{\beta_i}{K}\mathbf 1_K\mathbf 1_K^\top$ with $\beta_i\in[0,1]$.
Letting $\alpha_i:=1-\beta_i$ and $\bar\alpha_i:=\prod_{s=1}^{i}\alpha_s$, the cumulative matrix is
$\overline Q_i=\bar\alpha_i I+\frac{1-\bar\alpha_i}{K}\mathbf 1_K\mathbf 1_K^\top$~\cite{austin2021structured}.
The schedule is given by $\bar\alpha_i=a(i/M)$, where $a:[0,1]\to(0,1]$ is nonincreasing and $L$-Lipschitz,
with $a(0)=1$ and $a(\tau)<1$ for $\tau\in(0,1]$.
\end{assumption}

The uniform kernel~\cite{hoogeboom2021argmax, austin2021structured} in~\Cref{assum:schedule} is doubly stochastic, so the uniform distribution is stationary~\cite[A.1]{austin2021structured}. Using that, we define
\eqn{
\label{eq:uniform}
p_M\big(\xbf^{(M)}\big):=\operatorname{Cat}\!\Big(\xbf^{(M)};\,\tfrac{1}{K}\mathbf{1}_N\mathbf{1}_K^\top\Big).
}

\textbf{Reverse (denoising) process:} The reverse process starts from $\xbf^{(M)}\sim p_M$ and aims to recover a low-cost solution. At each step $i=M,\dots,1$, $\xbf^{(i-1)}$ is sampled from the learned reverse transition
\eqn{
\label{eq:combined}
p_{\theta}\big(&\xbf^{(i-1)}\mid \xbf^{(i)}\big)
= \sum_{\tilde \xbf^{(0)}}
q\big(\xbf^{(i-1)} \mid \xbf^{(i)}, \tilde \xbf^{(0)}\big)
p_{\theta}\big(\tilde \xbf^{(0)} \mid \xbf^{(i)}\big),
}
with the posterior distribution
% \seqn[\label{eq:posterior}]{
% q&\big(\xbf^{(i-1)} \mid  \xbf^{(i)}, \xbf^{(0)}\big)  
% =
% \frac {q\big(\xbf^{(i)} \mid \xbf^{(i-1)}\big)q\big(\xbf^{(i-1)} \mid \xbf^{(0)}\big)}{q\big(\xbf^{(i)} \mid \xbf^{(0)}\big)} \label{eq:posterior1}
% \\
% & = {\rm Cat}\left(\xbf^{(i-1)};
% \left( \left( \xbf^{(i)} \oslash (\xbf^{(0)} \overline{Q}_i) \right) Q_i^\top \right) \odot (\xbf^{(0)} \overline{Q}_{i-1}) \right)\label{eq:posterior2} 
% % \\ & = {\rm Cat}\left(\xbf^{(i-1)}; \rownorm\left(\xbf^{(i)} Q_i^\top \odot  \xbf^{(0)}\overline{Q}_{i-1}\right)\right) \label{eq:posterior3}
% }
\seqn[\label{eq:posterior}]{
q\big(\xbf^{(i-1)} &\mid  \xbf^{(i)}, \xbf^{(0)}\big)  
=
\frac {q\big(\xbf^{(i)} \mid \xbf^{(i-1)}\big)q\big(\xbf^{(i-1)} \mid \xbf^{(0)}\big)}{q\big(\xbf^{(i)} \mid \xbf^{(0)}\big)} \label{eq:posterior1}
\\
\label{eq:posterior2}
& = \prod_{r=0}^{N-1} {\rm Cat}\left(\xbf^{(i-1)}_r;\,
  \frac{\xbf^{(i)}_r Q_i^\top \odot \xbf^{(0)}_r\overline{Q}_{i-1}}
{\xbf^{(0)}_r\overline{Q}_i(\xbf^{(i)}_r)^\top}\right),
% \\  \label{eq:posterior3}
% & =:  {\rm Cat}\left(\xbf^{(i-1)}; \frac{\xbf^{(i)} Q_i^\top \odot  \xbf^{(0)}\overline{Q}_{i-1}}{\xbf^{(0)}\overline{Q}_i(\xbf^{(i)})^\top}\right) 
}
which follows from Bayes' theorem and the Markov property of the forward process.

Each factor in~\eqref{eq:posterior2} is the~\ac{D3PM} posterior~\cite[Eq.~(3)]{austin2021structured} applied to row $r$. Because this posterior is available in closed form,~\ac{D3PM} trains a neural network only to predict the clean-solution distribution $p_\theta\big(\tilde\xbf^{(0)}\mid\xbf^{(i)}\big)$. Then,~\eqref{eq:combined} is used to sample $\xbf^{(i-1)}$ for the next denoising step.

\section{Discrete Model-Based Diffusion}
\label{sec:method}

\begin{figure*}
    \centering
\includegraphics[width=\linewidth]{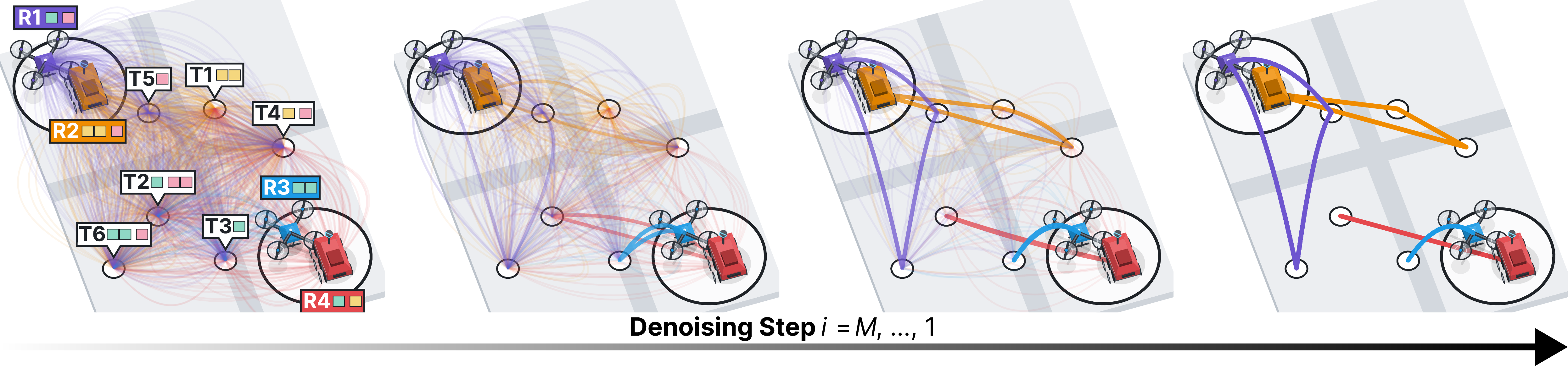}
    \caption{Visualization of denoising process in~\acf{DMBD} in the heterogeneous MRTA problem from~\Cref{fig:hero}. By performing diffusion directly over the discrete solution space of coalition formations and routes, the reverse transition distribution progressively concentrates its probability mass in solutions with shorter makespans, steering the search toward high-quality solutions.}
    \label{fig:discrete-mbd}
\end{figure*}

We propose~\acf{DMBD}, which shares the forward and reverse processes of~\ac{D3PM} but performs denoising in a \emph{training-free} manner. \ac{DMBD} differs from~\ac{D3PM} in two ways: (a) instead of learning the reverse transition $p(\xbf^{(i-1)}\mid \xbf^{(i)})$~\eqref{eq:combined} through offline training, it \textit{approximates} it with the target distribution~$p_0$~\eqref{eq:target_dist}, and (b) it aims to find a minimum-cost solution of~\eqref{eq:original_problem} rather than merely generate feasible solutions.

\subsection{\ac{DMBD} for Generic Optimization}
\label{sec:generic}

We first present~\ac{DMBD} for a generic combinatorial optimization problem. Its adaptation to heterogeneous MRTA problems is discussed in~\Cref{sec:robots}. 

\ac{DMBD}'s reverse process starts with sampling $\xbf^{(M)}\sim p_M$~\eqref{eq:uniform}. At each denoising step $i=M,\dots,1$,~\ac{DMBD} (a) samples candidate solutions from a proposal, (b) uses them to approximate $p\big(\xbf^{(i-1)}\mid\xbf^{(i)}\big)$, and (c) samples again from the approximated distribution and selects the best sample as $\xbf^{(i-1)}$.

\emph{(a) Sample.} By Bayes' theorem, we have 
$p(\xbf^{(0)}\mid\xbf^{(i)})\propto
q(\xbf^{(i)}\mid\xbf^{(0)})p_0(\xbf^{(0)})$. However, since $p_0$ couples all rows of $\xbf^{(0)}$, sampling from the normalized posterior requires summing over all $K^N$ elements of $\Xcal_{N,K}$, which is intractable. 

Instead, we construct a tractable proposal by examining the structure of the posterior. Note that (i) $q\big(\xbf^{(i)}\mid\xbf^{(0)}\big)$ factorizes over rows~\eqref{eq:i_step_forward}, with
$q\big(\xbf^{(i)}_r\mid\xbf^{(0)}_r\big)=\xbf^{(0)}_r\,\overline Q_i\big(\xbf^{(i)}_r\big)^\top=\xbf^{(0)}_r\big(\xbf^{(i)}_r\overline Q_i^\top\big)^\top$, and (ii) $\overline Q_i$ is doubly stochastic by~\Cref{assum:schedule}. By (i), the likelihood of each row $\xbf^{(0)}_r$ is given by $\xbf^{(i)}_r\overline Q_i^\top$, which by (ii) sums to one and thus defines a categorical distribution over $\xbf^{(0)}_r$. Hence, our proposal is 
\eqn{\label{eq:proposal}
\phi_i\big(\xbf^{(0)}\big) := \operatorname{Cat}\big(\xbf^{(0)};\,\xbf^{(i)}\overline Q_i^\top\big),
}
whose rows can be sampled independently. From the proposal $\phi_i$~\eqref{eq:proposal}, we then draw a batch $\Xcal_{S_1}$ of $S_1$ candidate solutions.

\emph{(b) Approximate.} By construction, $\phi_i\big(\xbf^{(0)}\big)=q\big(\xbf^{(i)}\mid\xbf^{(0)}\big)$ for all $\xbf^{(0)}\in\Xcal_{N,K}$. Substituting this into Bayes' theorem with the fact that $\phi_i$ is normalized, we get
\seqn[\label{eq:posterior_as_expectation}]{
p\big(\xbf^{(0)}\mid\xbf^{(i)}\big)
&= \frac{q\big(\xbf^{(i)}\mid\xbf^{(0)}\big)\,p_0(\xbf^{(0)})}
        {\sum_{\xbf\in\Xcal_{N,K}}q\big(\xbf^{(i)}\mid\xbf\big)p_0(\xbf)} \\
&= \frac{\phi_i(\xbf^{(0)})\,p_0(\xbf^{(0)})}
        {\Ebb_{\xbf\sim\phi_i}\big[p_0(\xbf)\big]}.
}

Substituting~\eqref{eq:posterior_as_expectation} into~\eqref{eq:combined} in place of $p_{\theta}\big(\tilde \xbf^{(0)} \mid \xbf^{(i)}\big)$ gives
\eqnN{
p_{i}\big(\xbf^{(i-1)} \mid &\xbf^{(i)}\big)
= \sum_{\tilde{\xbf}^{(0)}}
q(\xbf^{(i-1)}\mid\xbf^{(i)}, \tilde{\xbf}^{(0)})\,
p\big(\tilde{\xbf}^{(0)}\mid \xbf^{(i)}\big) \\
&= \frac{
\sum_{\tilde{\xbf}^{(0)}}
q(\xbf^{(i-1)}\mid\xbf^{(i)}, \tilde{\xbf}^{(0)})
\phi_i(\tilde{\xbf}^{(0)})p_0(\tilde{\xbf}^{(0)})
}{\Ebb_{\xbf\sim\phi_i}\big[p_0(\xbf)\big]} \\
&= \frac{
\Ebb_{\xbf\sim\phi_i}\big[\,
q(\xbf^{(i-1)}\mid\xbf^{(i)}, \xbf)\,p_0(\xbf)\,\big]
}{
\Ebb_{\xbf\sim\phi_i}\big[\,p_0(\xbf)\,\big]
},
}
which we approximate using the samples $\Xcal_{S_1}$ drawn from $\phi_i$:
\eqn{
p_i\big(\xbf^{(i-1)} \mid \xbf^{(i)}\big) &\approx
\frac{
\sum_{\xbf_s \in \Xcal_{S_1}}
q(\xbf^{(i-1)}\mid\xbf^{(i)}, \xbf_s)\,p_0(\xbf_s)
}{
\sum_{\xbf_s \in \Xcal_{S_1}} p_0(\xbf_s)} \nonumber\\ \label{eq:new_combined}
&=: \hat p_i\big(\xbf^{(i-1)} \mid \xbf^{(i)}\big).
}

\emph{(c) Select.} We draw $S_2$ candidate solutions $\Xcal_{S_2}$ from $\hat p_i$~\eqref{eq:new_combined}. Then we choose the lowest-cost candidate from $\Xcal_{S}:=\Xcal_{S_1}\cup \Xcal_{S_2}$ to get $\xbf^{(i-1)} \in \argmax{\xbf_s\in \Xcal_{S}} \  p_0(\xbf_s)$.

After $M$ denoising steps,~\ac{DMBD} returns $\xbf^{(0)}$ as its final solution. \Cref{alg:discrete_mbd} summarizes the procedure.

\begin{algorithm}[t]
\SetKwInOut{Inputs}{Inputs}
\SetKwInOut{Outputs}{Outputs}
\small
\caption{\acf{DMBD} for Generic Optimization}
\label{alg:discrete_mbd}

\Inputs{temperature $\lambda$, denoising steps $M$, sample counts $S_1$, $S_2$}
\Outputs{Final solution $\xbf^{(0)}$}

Initialize $\xbf^{(M)} \sim p_M(\xbf)= \operatorname{Cat}\!\Big(\cdot;\,\tfrac{1}{K}\mathbf{1}_N\mathbf{1}_K^\top\Big)$

\For{$i = M,\dots, 1$}{
Sample $\Xcal_{S_1}\sim \phi_i(\cdot)={\rm Cat}\Big(\cdot;
\xbf^{(i)}\overline Q_i^\top\Big)$

% Compute the posterior~\eqref{eq:posterior}:
% \eqnN{q(\xbf^{(i-1)}&\mid \xbf^{(i)}, \xbf^{(0)}) \\ &= \prod_{r=0}^{N-1} {\rm Cat}\left(\xbf^{(i-1)}_r;\,
%   \frac{\xbf^{(i)}_r Q_i^\top \odot \xbf^{(0)}_r\overline{Q}_{i-1}}
% {\xbf^{(0)}_r\overline{Q}_i(\xbf^{(i)}_r)^\top}\right)}

Update the distribution~\eqref{eq:new_combined}:
\eqnN{\hat p_i(\xbf^{(i-1)}\mid \xbf^{(i)}) =
\frac{
\sum_{\xbf_s \in \Xcal_{S_1}}
q(\xbf^{(i-1)}\mid\xbf^{(i)}, \xbf_s) p_0(\xbf_s)
}{\sum_{\xbf_s \in \Xcal_{S_1}} p_0(\xbf_s)}.
}

Sample $\Xcal_{S_2} \sim \hat p_i(\cdot\mid \xbf^{(i)})$

Select $\xbf^{(i-1)} \in \argmax{\xbf_s \in \Xcal_{S_1}\cup \Xcal_{S_2}} \ p_0(\xbf_s)$
}

\Return{$\xbf^{(0)}$}.
\end{algorithm}

\textbf{Similarity with~\ac{MBD}:}
\ac{DMBD} builds on the structure of~\ac{D3PM} in the same way that~\ac{MBD}~\cite{pan2024model} builds on~\ac{DDPM}. In particular, the proposal~\eqref{eq:proposal} is the categorical counterpart of~\cite[Eq.~(8)]{pan2024model}, and the approximation~\eqref{eq:new_combined} parallels the Monte Carlo score estimate in~\cite[Eq.~(9)]{pan2024model}.

\textbf{Distinctions from~\ac{D3PM}:} Compared to~\ac{D3PM}, our method has two key distinctions. First, rather than learning the reverse transition distribution,~\ac{DMBD} approximates it with samples, whose costs are evaluated directly by~\eqref{eq:target_dist}. Many extensions of~\ac{D3PM} also exploit problem knowledge, but only on top of a learned denoiser, either as guidance~\cite{li2024fast,lei2026boosting,li2025derivative} or as the training objective~\cite{sanokowski2024diffusion, sanokowski2025scalable}.

Second, while~\ac{D3PM} produces a single intermediate solution $\xbf^{(i)}$ at the end of each denoising step $i$,~\ac{DMBD} generates multiple candidate solutions and selects the lowest-cost one. This reflects their different goals:~\ac{D3PM} primarily focuses on the \emph{generative} aspect of modeling the target distribution, whereas~\ac{DMBD} aims to find an \emph{optimal} solution. Accordingly,~\ac{DMBD} greedily selects the best solution at each denoising step to steer the diffusion process toward low-cost regions. A similar selection appears in~\cite{li2025derivative}, but it again relies on a pretrained model and/or a learned or approximated value function.

\textbf{Role of $Q_i$:}
Similar to~\ac{D3PM}, the transition matrices $Q_i$ govern how~\ac{DMBD} explores $\Xcal_{N,K}$. Under~\Cref{assum:schedule}, each row of $\xbf'\sim\phi_i$ stays the same with probability $\bar\alpha_i+(1-\bar\alpha_i)/K$, and otherwise switches to one of the other categories uniformly at random. In the early stage of the algorithm (large $i$), $\bar\alpha_i$ remains relatively small, making $\phi_i$ closer to uniform and inducing exploration. As $i$ decreases, $\bar\alpha_i\to 1$ and $\phi_i$ places more probability mass around $\xbf^{(i)}$, performing local refinement. The schedule $a$ thus controls the exploration-exploitation trade-off, which we analyze formally in the next subsection.

\subsection{Theoretical Analysis}
\label{sec:analysis}

Here we establish a probabilistic convergence guarantee for our method toward the set of global minimizers $\Xcal^\star\subseteq \Ccal$.

\begin{theorem}\label{thm:convergence}
Let Assumption~\ref{assum:schedule} hold, $\Ccal \neq \emptyset$, and $S_1\geq S_2\geq 1$.
Define
$\gamma_N:=\int_0^1(1-a(\tau))^N\,d\tau$, $\rho:=N^2L/a_{\min}$, where $a_{\min}:=a(1)$, and $[\cdot]_+:=\max\{0,\cdot\}$. Then, for every $M\in \Z_{>0}$ and $\lambda>0$,
\eqn{\label{eq:bound}
\Pr\big(\xbf^{(0)}\in \Xcal^\star\big)\ge\left(1-\Big(1-\frac{\gamma_N}{K^N}\Big)^{S_1M}\right)\left[1-\frac{\rho^{S_2}}{M^{S_2-1}}\right]_+.}
\end{theorem}
\begin{proof}
The proof is given in \Cref{sec:appendix}.
\end{proof}

\textbf{Exploration-exploitation trade-off:} The first term on the right-hand side of~\eqref{eq:bound} lower-bounds the probability of sampling a minimizer in $\Xcal^\star$ at least once, and the second term lower-bounds the probability of remaining in $\Xcal^\star$ until the end of the algorithm. Together, they characterize the exploration-exploitation trade-off. A faster-decaying noise schedule $a$ keeps the noise level $1-\bar\alpha_i$ high for longer, which increases $\gamma_N$ and the first term. This matches the intuition that higher noise makes $\phi_i$ closer to uniform and promotes exploration. At the same time, it requires a larger $L$, which increases $\rho$ and shrinks the second term, consistent with the intuition that more exploration weakens exploitation. A similar trade-off for diffusion-based methods is discussed in~\cite{yi2026global}.

\textbf{Effects of sample sizes and diffusion steps:} 
Based on~\Cref{thm:convergence}, increasing the sample sizes $S_1$ and $S_2$ or the number of denoising steps $M$ raises the guaranteed probability of finding an optimal solution whenever the bound is nontrivial, i.e., $M^{S_2-1}>\rho^{S_2}$. More importantly, as $M\to\infty$, this lower bound monotonically increases for any $S_1\ge S_2\ge2$ and $\lambda>0$. This is consistent with the empirical results from~\cite{sun2023difusco,sanokowski2024diffusion}, where more denoising steps yield better solutions.

\section{\ac{DMBD} for Heterogeneous MRTA}
\label{sec:robots}

\textbf{Problem setup:} We now apply~\ac{DMBD} to the heterogeneous MRTA problem of~\Cref{sec:intro} and~\Cref{fig:hero}, an XD~[ST-MR-TA] instance in the taxonomy of~\cite{korsah2013comprehensive}. Consider a team of $R$ heterogeneous robots $\Rcal=\{0,\dots,R-1\}$ and $T$ spatially distributed tasks $\Tcal=\{0,\dots,T-1\}$ in the domain $\Dcal:=[0,1]^2$, with $B$ skill types $\Bcal=\{0,\dots,B-1\}$. We adopt the species-traits model~\cite{fu2023robust,dai2025heterogeneous}, where each robot belongs to one of the $S$ species, each with a unique set of skills. Robot $r$ has a capability vector $\mathbf{c}_r\in\Z_{\ge0}^B$, whose $b$-th entry is its capability level in skill $b$, a depot $\mathbf{o}_r\in\Dcal$, and a speed $v_r>0$, so that it travels between $\mathbf{z},\mathbf{z}'\in\Dcal$ in time $\Delta_r(\mathbf{z},\mathbf{z}'):=\|\mathbf{z}-\mathbf{z}'\|_2/v_r$. Task $j$ has a location $\mathbf{p}_j\in\Dcal$, a duration $\delta_j>0$, and a nonzero combined capability requirement vector for skills $\mathbf{w}_j\in\Z_{\ge0}^B$, whose $b$-th entry is the required minimum combined capability in skill $b$.

A solution consists of a coalition $\Lcal_j\subseteq\Rcal$ and a start time $t_j\ge0$ for each task $j$, and a route $\pi_r=(\pi_{r,0},\dots,\pi_{r,m_r-1})$ for each robot $r$, which lists the $m_r$ tasks $\{j\in\Tcal: r\in\Lcal_j\}$ assigned to robot $r$ in visiting order, starting and ending at its depot $\mathbf{o}_r$. Under tight coordination, task $j$ starts only once all members of $\Lcal_j$ have arrived, and all members remain at $\mathbf{p}_j$ until $t_j+\delta_j$. Minimizing the makespan $C_{\max}$, i.e., the time at which the last robot returns to its depot, yields
\seqn[\label{eq:mrta}]{
&\min_{\{\Lcal_j\},\{\pi_r\},\{t_j\},C_{\max}}\  C_{\max} \\
\text{s.t.}\ & \textstyle\sum_{r\in\Lcal_j}\mathbf{c}_r \succeq \mathbf{w}_j, \quad \forall j\in\Tcal, \label{eq:mrta_skill}\\
& t_{\pi_{r,0}} \ge \Delta_r\big(\mathbf{o}_r,\mathbf{p}_{\pi_{r,0}}\big), \quad \forall r\in\Rcal: m_r>0, \label{eq:routing_constraint1}    \\
& t_{\pi_{r,\ell+1}} \ge t_{\pi_{r,\ell}}+\delta_{\pi_{r,\ell}}+\Delta_r\big(\mathbf{p}_{\pi_{r,\ell}},\mathbf{p}_{\pi_{r,\ell+1}}\big), \nonumber\\
& \qquad\qquad \forall r\in\Rcal,\ \ell=0,\dots,m_r-2,\\
& C_{\max} \ge t_{\pi_{r,m_r-1}}+\delta_{\pi_{r,m_r-1}}+\Delta_r\big(\mathbf{p}_{\pi_{r,m_r-1}},\mathbf{o}_r\big), \nonumber\\
& \qquad\qquad \forall r\in\Rcal: m_r>0.   \label{eq:routing_constraint2}
}

We assume problem~\eqref{eq:mrta} itself is feasible, i.e., $\sum_{r\in\Rcal}\mathbf{c}_r\succeq\mathbf{w}_j$ for all $j\in\Tcal$. 

We study three variants, SS-BT, SS-AT, and MS-AT, which differ in robot skills and task requirements. In SS (single-skill agents), robots have one skill, i.e., $\mathbf{c}_r$ has exactly one nonzero entry, whereas in MS (multi-skill agents), robots may have several skills. In BT (binary tasks), $\mathbf{w}_j\in\{0,1\}^B$, so one capable robot per required skill satisfies~\eqref{eq:mrta_skill}. In AT (additive tasks), requirements are general nonnegative integers that the coalition must meet through its combined capabilities.

\textbf{Priority-based encoding:}
\ac{DMBD} could be directly applied to~\eqref{eq:mrta}, but most of its samples would violate the constraints, severely reducing sample efficiency. Thus, following~\cite{bean1994genetic,storer1992new}, we instead represent candidate solutions as robot-task priorities that can be decoded into feasible solutions for~\eqref{eq:mrta}. The decision variable is $\xbf\in\Xcal_{RT,T}$, where each row indexed by $(r,j)\in\Rcal\times\Tcal$ specifies the priority level robot $r$ assigns to task $j$. The entry $[\xbf]_{(r,j)k}=1$ indicates priority level $k$, with smaller $k$ denoting higher priority. We evaluate each candidate $\xbf$ by decoding it into a solution as follows.

\textbf{Decoding:}
Given $\xbf\in\Xcal_{RT,T}$, let $\kappa_{rj}(\xbf)\in\{0,\dots,T-1\}$ denote the priority level that robot $r$ assigns to task $j$, i.e., $[\xbf]_{(r,j)\kappa_{rj}(\xbf)}=1$. We then evaluate $\xbf$ on~\eqref{eq:mrta} as follows.

% , and let $\Rcal_b:=\{r\in\Rcal : [\mathbf{c}_r]_b\geq1\}$ be the set of robots with skill $b$

\emph{(a) Coalition formation.} For each task $j\in\Tcal$, visit the robots from highest to lowest priority for task $j$ (i.e., increasing $\kappa_{rj}(\xbf)$, breaking ties by robot index). Add robot $r$ to $\Lcal_j(\xbf)$ only if doing so reduces the unmet requirement $\sum_{b\in\Bcal}\max\{[\mathbf{w}_j]_b - \sum_{r'\in\Lcal_j(\xbf)}[\mathbf{c}_{r'}]_b,\,0\}$. 

\emph{(b) Coalition pruning (only for MS-AT).} Since robots may possess several skills in MS-AT, $\Lcal_j(\xbf)$ can contain a larger team than necessary. For each task $j\in\Tcal$, visit the members of $\Lcal_j(\xbf)$ from lowest to highest priority (breaking ties by robot index) and remove robot $r$ whenever the remaining coalition still meets the requirement, i.e., $\sum_{r'\in\Lcal_j(\xbf)\setminus\{r\}}\mathbf{c}_{r'} \succeq \mathbf{w}_j$.

\emph{(c) Global task ordering.} To enforce a global order over tasks, each robot visits its assigned tasks in order of their coalition’s average priority, $s_j(\xbf) := \frac{1}{|\Lcal_j(\xbf)|}\sum_{r\in\Lcal_j(\xbf)}\kappa_{rj}(\xbf)$, from highest to lowest priority (i.e., ascending $s_j$), breaking ties by task index.

\emph{(d) Scheduling.}
Simulate the resulting routes, starting each task as soon as all coalition members arrive. This yields the start and end times of all tasks and the makespan as $J(\xbf)=C_{\max}$.

These steps decode every $\xbf\in\Xcal_{RT,T}$ into a feasible solution of~\eqref{eq:mrta}. Steps~(a)-(b) select coalitions that satisfy~\eqref{eq:mrta_skill}. Step~(c) prevents cyclic waiting among coalitions, satisfying the routing constraints~\eqref{eq:routing_constraint1}-\eqref{eq:routing_constraint2}. Hence,~\ac{DMBD} applies directly with $p_0(\xbf)\propto\exp(-J(\xbf)/\lambda)$.

% \textbf{Plan construction for MA-AT:}
% When robots possess multiple skills, the coalition-formation procedure above does not necessarily satisfy~\eqref{eq:mrta_skill}, since selecting robots separately for each skill may lead to overlapping skill contributions. We therefore retain the same priority-based representation and plan-construction procedure but penalize unmet task requirements during evaluation. Specifically, we optimize
% \eqn{
% \label{eq:penalty}
% J_\mu(\xbf) := J(\xbf) + \mu\sum_{j\in\Tcal}\Big\|\Big[\mathbf{w}_j-\textstyle\sum_{r\in\Lcal_j(\xbf)}\mathbf{c}_r\Big]_+\Big\|_1.
% }
% where $[\cdot]_+$ denotes the componentwise positive part and $\mu>0$ is the penalty weight. The penalty is zero when all skill requirements are satisfied and strictly positive whenever any requirement is unmet. Since $J\ge0$ and any violation contributes at least $\mu$, having a large $\mu$ encourages a selection of a feasible $\xbf$ as every minimizer of $J_\mu$ is feasible.

\begin{remark}
Our priority-based encoding improves the sample efficiency of~\ac{DMBD} by ensuring that every sample decodes into a feasible solution. However, the encoding is neither injective nor surjective: different priority assignments may be decoded to the same plan, and some solutions of~\eqref{eq:mrta} may not be representable. A common alternative is penalty-based constraint handling, which penalizes infeasible solutions. More general approaches to constraint handling while maintaining sample efficiency for~\ac{DMBD} remain future work.
\end{remark}

\section{Simulation Results}
We evaluate~\ac{DMBD} on generic, unconstrained integer programs and on heterogeneous MRTA, addressing two questions: \textbf{Q1.} How does~\ac{DMBD} compare with other SBO methods in solution quality and computation time? \textbf{Q2.} How does it compare with exact, metaheuristic, and learning-based methods? 
% \textbf{Q3.} Does the convergence predicted by~\Cref{thm:convergence} hold in practice, i.e., how does the number of denoising steps $M$ affect optimality?

\subsection{Experimental Setup}\label{sec:setup}
\textbf{Metrics:} We report the mean objective value (cost for generic integer programs, makespan for MRTA) and the computation time per instance, averaged over 5 solver seeds and all problem instances. Computation times exclude one-time JIT compilation. Averages include only runs in which the solver returns a feasible solution.

\textbf{Implementation:} All sampling-based methods are implemented in Python using JAX~\cite{jax2018github}, with sample evaluations parallelized on an NVIDIA RTX 3080 Ti GPU, while Gurobi~\cite{gurobi} runs on the CPU. All experiments are conducted on a computer equipped with an Intel Core i9-12900KF CPU and 64 GB of RAM. For~\ac{DMBD}, we use temperature $\lambda=10^{-5}$ and the uniform transition matrix~\cite{hoogeboom2021argmax,austin2021structured} with the modified cosine schedule:
\eqnN{\bar\alpha_i = a_{\min} + (1-a_{\min})\,\frac{f(i/M)}{f(0)},\ \
f(\tau)=\cos^2\!\Big(\frac{\pi}{2}\cdot\frac{\tau+s}{1+s}\Big), }
where $s=0.008$ and $a_{\min}=10^{-3}$, thereby satisfying~\Cref{assum:schedule}.
At each denoising step, we allocate $75\%$ of the samples to $S_1$ and the remaining $25\%$ to $S_2$.

\subsection{Unconstrained Problems}
\label{sec:sim-unconstrained}

% Requires \usepackage{booktabs,multirow,graphicx,colortbl}
\begin{table*}[t]
\centering
\setlength{\tabcolsep}{2.5pt}
\caption{Mean cost and runtime (s) on unconstrained UIQP and UPO over 50 instances across 5 solver seeds. Costs share the scale factor shown; \textbf{bold}/\underline{underline} mark the best/second-best method besides Gurobi with 30 s limit ($^\dagger$: Gurobi hit its 30\,s time limit).}
\label{tab:results_unconstrained}
\resizebox{\textwidth}{!}{%
\begin{tabular}{llcccccc|cccccc}
\toprule
 & & \multicolumn{6}{c|}{\textbf{UIQP}} & \multicolumn{6}{c}{\textbf{UPO}} \\
\cmidrule(lr){3-8}\cmidrule(lr){9-14}
 & & \multicolumn{2}{c}{$K=10$} & \multicolumn{2}{c}{$K=50$} & \multicolumn{2}{c|}{$K=100$} & \multicolumn{2}{c}{$K=10$} & \multicolumn{2}{c}{$K=50$} & \multicolumn{2}{c}{$K=100$} \\
\cmidrule(lr){3-4}\cmidrule(lr){5-6}\cmidrule(lr){7-8}\cmidrule(lr){9-10}\cmidrule(lr){11-12}\cmidrule(lr){13-14}
 & Method & Cost $\downarrow$ & Time (s) $\downarrow$ & Cost $\downarrow$ & Time (s) $\downarrow$ & Cost $\downarrow$ & Time (s) $\downarrow$ & Cost $\downarrow$ & Time (s) $\downarrow$ & Cost $\downarrow$ & Time (s) $\downarrow$ & Cost $\downarrow$ & Time (s) $\downarrow$ \\
\midrule
\multirow{6}{*}{$N=10$} & Gurobi (30 s) & $-10.01$ & 0.11 & $-2.95$ & 0.19 & $-1.20$ & 0.51 & $-10.01$ & 22.16 & $-4.72$ & 30$^{\dagger}$ & $-1.57$ & 30$^{\dagger}$ \\ \cmidrule(lr){2-14}
 & CEM & $-9.78$ & 0.47 & $-2.79$ & 0.47 & $-1.13$ & 0.47 & \underline{$-9.71$} & 0.48 & $-4.37$ & 0.48 & $-1.45$ & 0.49 \\
 & SMC & $\mathbf{-9.85}$ & 0.86 & \underline{$-2.89$} & 0.86 & \underline{$-1.18$} & 0.86 & $-9.59$ & 0.86 & \underline{$-4.61$} & 0.86 & \underline{$-1.55$} & 0.86 \\
 & Rounded MBD & $-9.73$ & 0.47 & $-2.48$ & 0.47 & $-0.93$ & 0.56 & $-9.22$ & 0.47 & $-3.03$ & 0.46 & $-0.81$ & 0.56 \\
\rowcolor[gray]{0.93}
 & Discrete MBD (ours) & \underline{$-9.84$} & 0.56 & $\mathbf{-2.89}$ & 0.58 & $\mathbf{-1.18}$ & 0.61 & $\mathbf{-9.80}$ & 0.57 & $\mathbf{-4.67}$ & 0.58 & $\mathbf{-1.57}$ & 0.61 \\
\cmidrule(lr){2-14}
 & Scale & $10^{2}$ & -- & $10^{4}$ & -- & $10^{5}$ & -- & $10^{5}$ & -- & $10^{9}$ & -- & $10^{11}$ & -- \\
\midrule
\multirow{6}{*}{$N=50$} & Gurobi (30 s) & $-1.17$ & 27.24 & $-3.46$ & 30$^{\dagger}$ & $-1.41$ & 30$^{\dagger}$ & $-2.28$ & 30$^{\dagger}$ & $-1.08$ & 30$^{\dagger}$ & $-3.69$ & 30$^{\dagger}$ \\ \cmidrule(lr){2-14}
 & CEM & $-1.14$ & 0.47 & $-3.16$ & 0.48 & $-1.28$ & 0.49 & $-2.14$ & 0.48 & $-0.90$ & 0.49 & $-2.95$ & 0.50 \\
 & SMC & \underline{$-1.18$} & 0.87 & \underline{$-3.27$} & 0.87 & \underline{$-1.32$} & 0.87 & \underline{$-2.28$} & 0.86 & \underline{$-0.96$} & 0.86 & \underline{$-3.14$} & 0.86 \\
 & Rounded MBD & $-0.82$ & 0.45 & $-1.67$ & 0.48 & $-0.52$ & 0.44 & $-1.11$ & 0.45 & $-0.23$ & 0.48 & $-0.40$ & 0.44 \\
\rowcolor[gray]{0.93}
 & Discrete MBD (ours) & $\mathbf{-1.20}$ & 0.60 & $\mathbf{-3.54}$ & 0.59 & $\mathbf{-1.43}$ & 0.60 & $\mathbf{-2.36}$ & 0.60 & $\mathbf{-1.12}$ & 0.59 & $\mathbf{-3.71}$ & 0.60 \\
\cmidrule(lr){2-14}
 & Scale & $10^{4}$ & -- & $10^{5}$ & -- & $10^{6}$ & -- & $10^{6}$ & -- & $10^{10}$ & -- & $10^{11}$ & -- \\
\midrule
\multirow{6}{*}{$N=100$} & Gurobi (30 s) & $-3.30$ & 30$^{\dagger}$ & $-0.98$ & 30$^{\dagger}$ & $-4.00$ & 30$^{\dagger}$ & $-3.24$ & 30$^{\dagger}$ & $-1.56$ & 30$^{\dagger}$ & $-5.22$ & 30$^{\dagger}$ \\ \cmidrule(lr){2-14}
 & CEM & \underline{$-3.04$} & 0.47 & \underline{$-0.83$} & 0.49 & \underline{$-3.35$} & 0.50 & \underline{$-2.68$} & 0.48 & \underline{$-1.11$} & 0.51 & \underline{$-3.61$} & 0.52 \\
 & SMC & $-2.80$ & 0.92 & $-0.75$ & 0.93 & $-3.05$ & 0.93 & $-2.37$ & 0.92 & $-0.95$ & 0.93 & $-3.14$ & 0.93 \\
 & Rounded MBD & $-1.73$ & 0.45 & $-0.32$ & 0.44 & $-0.97$ & 0.47 & $-0.99$ & 0.45 & $-0.19$ & 0.44 & $-0.32$ & 0.47 \\
\rowcolor[gray]{0.93}
 & Discrete MBD (ours) & $\mathbf{-3.33}$ & 0.59 & $\mathbf{-0.93}$ & 0.58 & $\mathbf{-3.76}$ & 0.58 & $\mathbf{-3.23}$ & 0.59 & $\mathbf{-1.39}$ & 0.58 & $\mathbf{-4.59}$ & 0.58 \\
\cmidrule(lr){2-14}
 & Scale & $10^{4}$ & -- & $10^{6}$ & -- & $10^{6}$ & -- & $10^{6}$ & -- & $10^{10}$ & -- & $10^{11}$ & -- \\
\bottomrule
\end{tabular}%
}
\end{table*}

\textbf{Settings:} We consider two classes of unconstrained problems over $N$ variables with $K$ categories each. Both are written in terms of the integer vector $\ybf\in\{0,\dots,K-1\}^N$, with $[\ybf]_r := \sum_{k=0}^{K-1} k\,[\xbf]_{rk}$. For each problem class and pair $N,K\in\{10,50,100\}$, we evaluate methods on 50 random instances.

\textit{Unconstrained Integer Quadratic Program (UIQP):}
\eqn{
\label{eq:uiqp}
\min_{\ybf \in \{0,\ldots,K-1\}^N}
\quad
\ybf^\top  \mathbf{R}\,\ybf + \mathbf{h}^\top\ybf ,
}
where $\mathbf{R}\in\R^{N\times N}$ is symmetric and indefinite and $\mathbf{h}\in\R^N$. For $K=2$, \eqref{eq:uiqp} reduces to quadratic unconstrained binary optimization (QUBO), which is equivalent to an Ising problem~\cite{lucas2014ising}. We set $\mathbf{R}:=\frac 1 2(\mathbf{R}'+(\mathbf{R}')^\top)$ with $[\mathbf{R}']_{ij}\sim\mathcal{N}(0,1)$ i.i.d., and $\mathbf{h}\sim\mathcal{N}(\mathbf{0},I)$.

\textit{Unconstrained Polynomial Optimization (UPO):}
\eqn{
\label{eq:upo}
\min_{\ybf \in \{0,\ldots,K-1\}^N}
\quad
\sum_{u=1}^{U}
\mu_u
\prod_{r \in \mathcal{P}_u} \ybf_r ,
}
where monomial $u$ has coefficient $\mu_u\in\R$ and index set $\Pcal_u\subseteq\{0,\dots,N-1\}$. For $K=2$, \eqref{eq:upo} is a pseudo-Boolean optimization problem~\cite{boros2002pseudo}. We use $U=250$ monomials, each over $|\Pcal_u|=5$ distinct variables chosen uniformly at random, which keeps the objective magnitude moderate and ensures numerical stability.

% \textbf{Baselines:} We compare against the cross-entropy method (CEM)~\cite{rubinstein1999cross} with elite fraction $1\%$; simulated annealing (SA)~\cite{bertsimas1993simulated} with initial temperature $1.0$ and cooling factor $0.95$; sequential Monte Carlo (SMC)~\cite{del2006sequential}, which mutates each particle by resampling $10\%$ of the variables; and large neighborhood search (LNS)~\cite{pisinger2018large}, which destroys $10\%$ of the variables and repairs them with $k$-exchange neighborhoods. We also evaluate Rounded~\ac{MBD}, which runs~\ac{MBD} on the $[-1,1]^{N\times K}$ relaxation with the cosine noise schedule~\cite{nichol2021improved}, maps each sample to $[0,1]^{N\times K}$ via $\tfrac12(\mathbf 1_N\mathbf 1_K^\top+\xbf)$, and rounds it to a one-hot configuration as in~\cite{sun2023difusco}. For~\ac{DMBD}, we use the uniform transition matrix (\Cref{assum:schedule}) with the cosine schedule of~\cite{hoogeboom2021argmax,austin2021structured} and allocate $75\%$ and $25\%$ of each step's samples to $S_1$ and $S_2$, respectively. Both diffusion methods use $\lambda=10^{-5}$. All algorithms run for $M=100$ iterations with $2048$ samples per iteration. Lastly, we include Gurobi~\cite{gurobi} with a 30\,s time limit, averaged over $5$ different initial states, as a reference for solution quality. 

\textbf{Baselines:} We consider three SBO methods: the cross-entropy method (CEM)~\cite{rubinstein1999cross} with an elite fraction of $1\%$; sequential Monte Carlo (SMC)~\cite{del2006sequential}, which mutates each particle by resampling $10\%$ of its variables; and Rounded~\ac{MBD}, which runs~\ac{MBD} with temperature $\lambda=10^{-5}$ on the relaxation $[-1,1]^{N\times K}$ with the cosine noise schedule~\cite{nichol2021improved}, and rounds each sample to a one-hot form via a row-wise argmax, as in~\cite{sun2023difusco}. For reference on solution quality only, we also report Gurobi~\cite{gurobi} with a 30\,s time limit, averaged over 5 initial states. All methods, including \ac{DMBD}, run for $M=100$ iterations with $2048$ samples per iteration.

\textbf{Results (Q1):} \Cref{tab:results_unconstrained} shows that~\ac{DMBD} achieves the lowest average cost among the SBO methods in 17 out of 18 settings. The only exception is the smallest size ($N=K=10$), where it is still within $0.1\%$ of the best. \ac{DMBD} takes about 0.6\,s per problem, slightly more than CEM and Rounded~\ac{MBD}. Overall, \ac{DMBD} offers the best solution quality among SBO methods at comparable computational cost.

\subsection{Multi-Robot Coalition and Scheduling}
\label{sec:sim-mrta}

% Preamble:
% \usepackage{booktabs}
% \usepackage{graphicx}
% \usepackage[table]{xcolor}

\begin{table*}[t]
\centering
\small
\setlength{\tabcolsep}{3pt}
\renewcommand{\arraystretch}{0.90}

\caption{Mean makespan and computation time on Heterogeneous MRTA
problem~\eqref{eq:mrta} over 50 problem instances across 5 solver seeds. 
CTAS-D superscripts indicate success rates. \textbf{Bold}/\underline{underline} indicate best/second-best makespan, respectively.
}
\label{tab:paper-suite-compact}

\resizebox{\textwidth}{!}{%
\begin{tabular}{l*{5}{c@{\hspace{2pt}}c}}
\toprule
& \multicolumn{2}{c}{$R{=}9,\ S{=}3$}
& \multicolumn{2}{c}{$R{=}15,\ S{=}5$}
& \multicolumn{2}{c}{$R{=}25,\ S{=}5$}
& \multicolumn{2}{c}{$R{=}25,\ S{=}5$}
& \multicolumn{2}{c}{$R{=}50,\ S{=}5$} \\
& \multicolumn{2}{c}{$T{=}20,\ B{=}3$}
& \multicolumn{2}{c}{$T{=}20,\ B{=}5$}
& \multicolumn{2}{c}{$T{=}20,\ B{=}5$}
& \multicolumn{2}{c}{$T{=}50,\ B{=}5$}
& \multicolumn{2}{c}{$T{=}50,\ B{=}5$} \\
\cmidrule(lr){2-3}\cmidrule(lr){4-5}\cmidrule(lr){6-7}\cmidrule(lr){8-9}\cmidrule(lr){10-11}
Method
& \footnotesize Makespan $\downarrow$ & \footnotesize Time (s) $\downarrow$
& \footnotesize Makespan $\downarrow$ & \footnotesize Time (s) $\downarrow$
& \footnotesize Makespan $\downarrow$ & \footnotesize Time (s) $\downarrow$
& \footnotesize Makespan $\downarrow$ & \footnotesize Time (s) $\downarrow$
& \footnotesize Makespan $\downarrow$ & \footnotesize Time (s) $\downarrow$ \\
\midrule

\multicolumn{11}{l}{\textit{SS-BT}} \\
\specialrule{0.4pt}{0.15em}{0.15em}
CTAS-D (30 s)
& 31.39\textsuperscript{\ 61\%} & 30
& 40.74\textsuperscript{\ 36\%} & 30
& 23.31\textsuperscript{\ 98\%} & 30
& -- & 30
& 104.15\textsuperscript{\ $<1\%$} & 30 \\
LNS
& 24.94 & 0.29
& 25.85 & 0.32
& 18.69 & 0.35
& 51.04 & 0.66
& 30.73 & 1.49 \\
SA
& 26.21 & 0.09
& 28.08 & 0.11
& 21.22 & 0.13
& 53.54 & 0.42
& 36.32 & 0.91 \\
RL(s.10)
& 27.30 & 1.61
& 26.93 & 2.58
& 19.96 & 3.22
& \textbf{35.44} & 7.08
& \underline{22.91} & 9.95 \\
\addlinespace[5pt]
CEM
& 26.82 & 0.08
& 27.76 & 0.11
& 20.32 & 0.12
& 44.95 & 0.50
& 27.48 & 1.09 \\
SMC
& \textbf{24.64} & 0.34
& \underline{25.51} & 0.37
& \underline{18.48} & 0.40
& 50.84 & 0.72
& 30.65 & 1.56 \\
Rounded MBD
& 27.53 & 0.28
& 29.88 & 0.19
& 21.79 & 0.29
& 60.67 & 1.44
& 37.52 & 3.10 \\
\rowcolor[gray]{0.93}
Discrete MBD (ours)
& \underline{24.88} & 0.15
& \textbf{25.25} & 0.17
& \textbf{17.95} & 0.20
& \underline{37.16} & 0.86
& \textbf{22.03} & 1.70 \\
\midrule
\addlinespace[5pt]
\multicolumn{11}{l}{\textit{SS-AT}} \\
\specialrule{0.4pt}{0.15em}{0.15em}
CTAS-D (30 s)
& -- & 30
& -- & 30
& -- & 30
& -- & 30
& -- & 30 \\
LNS
& 44.97 & 0.29
& 54.51 & 0.32
& 40.98 & 0.35
& 130.86 & 0.66
& 83.28 & 1.49 \\
SA
& 46.01 & 0.09
& 55.44 & 0.11
& 41.99 & 0.13
& 115.03 & 0.42
& 80.02 & 0.90 \\
RL(s.10)
& 50.46 & 2.53
& 57.55 & 4.26
& \textbf{33.59} & 5.19
& \textbf{70.95} & 12.45
& \textbf{37.70} & 16.08 \\
\addlinespace[6pt]
CEM
& 46.87 & 0.08
& 54.31 & 0.11
& 38.97 & 0.12
& 99.39 & 0.50
& 60.80 & 1.09 \\
SMC
& \underline{44.34} & 0.34
& \underline{53.96} & 0.37
& 40.10 & 0.40
& 130.57 & 0.72
& 82.84 & 1.56 \\
Rounded MBD
& 50.49 & 0.28
& 61.52 & 0.19
& 47.47 & 0.29
& 143.52 & 1.44
& 93.82 & 3.11 \\
\rowcolor[gray]{0.93}
Discrete MBD (ours)
& \textbf{44.02} & 0.16
& \textbf{50.79} & 0.17
& \underline{34.23} & 0.20
& \underline{83.25} & 0.86
& \underline{47.65} & 1.69 \\
\midrule
\addlinespace[5pt]
\multicolumn{11}{l}{\textit{MS-AT}} \\
\specialrule{0.4pt}{0.15em}{0.15em}
CTAS-D (30 s)
& 62.29\textsuperscript{\ 24\%} & 30
& 80.36\textsuperscript{\ 10\%} & 30
& 39.51\textsuperscript{\ 19\%} & 30
& 126.54\textsuperscript{\ $<1\%$} & 30
& 67.60\textsuperscript{\ 3\%} & 30 \\
LNS
& 36.32 & 0.46
& 32.72 & 0.56
& 22.51 & 0.71
& 68.91 & 1.52
& 40.66 & 3.31 \\
SA
& 39.30 & 0.17
& 36.61 & 0.26
& 26.96 & 0.40
& 74.28 & 1.19
& 48.28 & 2.52 \\
RL(s.10)
& 41.44 & 2.21
& 35.21 & 3.16
& 23.80 & 3.87
& \textbf{47.15} & 8.93
& \textbf{27.92} & 12.27 \\
\addlinespace[6pt]
CEM
& 36.22 & 0.17
& 32.28 & 0.26
& 22.36 & 0.40
& 52.18 & 1.31
& 31.16 & 2.77 \\
SMC
& \textbf{35.49} & 0.49
& \underline{32.19} & 0.60
& \underline{22.12} & 0.75
& 68.53 & 1.57
& 40.32 & 3.38 \\
Rounded MBD
& 40.85 & 0.38
& 37.87 & 0.36
& 26.86 & 0.57
& 82.20 & 2.17
& 49.61 & 4.47 \\
\rowcolor[gray]{0.93}
Discrete MBD (ours)
& \underline{35.67} & 0.26
& \textbf{31.68} & 0.35
& \textbf{21.30} & 0.65
& \underline{49.67} & 1.82
& \underline{28.36} & 3.65 \\

\bottomrule
\end{tabular}%
}
\end{table*}

\textbf{Settings:} We now consider the heterogeneous MRTA problem~\eqref{eq:mrta}, using the benchmark environment and test sets of~\cite{dai2025heterogeneous}. Following~\cite{dai2025heterogeneous}, we consider five configurations of $(R,S,T,B)$ (\Cref{tab:paper-suite-compact}) for each of the three variants SS-BT, SS-AT, and MS-AT (called SA-BT, SA-AT, and MA-AT therein), giving 15 settings with 50 instances each. A run is successful if it returns a deadlock-free plan that completes every task.

% \textbf{Baselines:} We reuse the baselines from \Cref{sec:sim-unconstrained} with the same hyperparameters. In addition, we add two metaheuristics: large neighborhood search (LNS)~\cite{pisinger2018large}, which destroys $10\%$ of the variables and repairs them with $k$-exchange neighborhoods and simulated annealing (SA)~\cite{bertsimas1993simulated} with initial temperature $1.0$ and cooling factor $0.95$. Like~\ac{DMBD}, all of them use the priority-based representation and plan construction of \Cref{sec:robots}, so they differ only in how they search over $\Xcal_{RT,T}$. Also, we increase the number of diffusion steps to $M=350$ and fix the sample size to $2048$ for all methods. We additionally compare against one learning-based: RL(s.10)~\cite{dai2025heterogeneous}, a reinforcement-learning policy trained on MA-AT instances; it runs each instance 10 times with Boltzmann action selection and returns the plan with the lowest makespan. Then, we also include exact method: CTAS-D~\cite{fu2023robust} solves a mixed-integer linear program with Gurobi~\cite{gurobi}, first optimizing the flow of each species over the tasks and then rounding the flows to integers to obtain robot routes; we run the method with a 30\,s time limit per instance. 

\textbf{Baselines:} We group the baselines into four categories.
\textit{(i) MILP-based:} CTAS-D~\cite{fu2023robust}, which solves a mixed-integer linear program with Gurobi~\cite{gurobi} to optimize the flow of each species over the tasks, then rounds the flows to integers to obtain robot routes. We use a 30\,s time limit per instance.
\textit{(ii) Learning-based:} RL(s.10)~\cite{dai2025heterogeneous}, a reinforcement-learning policy trained on MS-AT instances. It runs each instance 10 times with Boltzmann action selection and returns the plan with the lowest makespan.
\textit{(iii) Metaheuristics:} large neighborhood search (LNS)~\cite{pisinger2018large}, which destroys $10\%$ of the variables and repairs them with $k$-exchange neighborhoods, and simulated annealing (SA)~\cite{bertsimas1993simulated} with initial temperature $1.0$ and cooling factor $0.95$.
\textit{(iv) SBO:} CEM, SMC, and Rounded~\ac{MBD} from \Cref{sec:sim-unconstrained}, with the same hyperparameters. All sampling-based baselines (metaheuristics and SBO) and~\ac{DMBD} use the encoding scheme described in~\Cref{sec:robots} and are run for $M=350$ iterations, with $2048$ samples per iteration.

\textbf{Results (Q1 \& Q2):} As shown in~\Cref{tab:paper-suite-compact},~\ac{DMBD} achieves the lowest makespan among the SBO methods in 13 out of 15 configurations, lagging behind only on the smallest instances ($R=9$). In fact, its performance advantage also widens as the problem size grows. Against all baselines, \ac{DMBD} achieves the lowest makespan in 7 of 15 settings and the second lowest in the remaining 8. RL(s.10) performs better on larger configurations, where~\ac{DMBD}, like other sampling-based methods, suffers from lower sample efficiency. Unlike RL(s.10), however,~\ac{DMBD} is training-free while still finding competitive solutions within $3.7$\,s. CTAS-D fails to find feasible solutions within 30\,s on many instances (as shown in the superscripts indicating the success rates), making it unsuitable for online planning. Overall, \ac{DMBD} outperforms all baselines or closely approaches the strongest baseline in every setting while remaining training-free and fast enough for online use.

% \subsection{Ablation Study}
% Finally, we empirically validate~\Cref{thm:convergence} through an ablation study on the number of denoising steps, $M$. According to~\Cref{thm:convergence}, increasing $M$ monotonically improves the lower bound on the convergence probability for any $S_1,S_2\geq 2$ and $\lambda>0$. Accordingly, we evaluate~\ac{DMBD} with $M\in\{10^{p}\}_{p=1}^6$ on 100 randomly generated instances of UPO with $N=50$ and $K=10$.

% \textbf{Results~(Q3):} \Cref{fig:ablation} shows the results, where as $M$ increases, the performance of~\ac{DMBD} improves. 

% \begin{figure}
%     \centering
% \includegraphics[width=\linewidth]{}
%     \caption{Caption}
%     \label{fig:placeholder}
% \end{figure}

\section{Conclusions}
\label{sec:conc}

In this paper, we proposed \acf{DMBD}, a training-free sampling-based optimization (SBO) method that replaces the learned denoiser of discrete diffusion with a Monte Carlo approximation and guarantees asymptotic convergence to the global optimum. We also presented a priority-based encoding for heterogeneous MRTA that guarantees feasible samples and thus improves sample efficiency. \ac{DMBD} outperforms SBO and metaheuristic baselines in nearly all settings and matches mixed-integer linear program-based and learning-based methods at a fraction of their computational cost.

Several directions remain for future work. First, \ac{DMBD} needs enough denoising steps to be competitive; with few steps, it may perform no better, or even worse, than other methods. Our convergence bound characterizes the role of the number of steps but becomes loose as the problem size grows. Using adaptive noise schedules to reduce the number of denoising steps would be an interesting direction. Second, like other SBO methods, \ac{DMBD} loses sample efficiency on larger problems. For heterogeneous MRTA, extending \ac{DMBD} to a distributed setting, as in~\cite{lee2026distributed}, is a promising direction. Lastly, we assume a deterministic cost $J$. Extending \ac{DMBD} to stochastic objectives would allow our MRTA formulation to capture uncertain skill requirements and travel times.
\section{Appendix}
\label{sec:appendix}

Before the proof for~\Cref{thm:convergence}, we first provide two supporting lemmas.

\begin{lemma}\label{lem:proposal}
Let~\Cref{assum:schedule} hold, and let $\xbf^{(i)}\in \Xcal_{N,K}$ be the solution at diffusion step $1\le i\le M$. Then, we have (A) $\phi_i(\xbf)\ge(\frac {1-\bar\alpha_i} K)^N$ for all $\xbf\in\Xcal_{N,K}$ and (B) $\phi_i(\xbf^{(i)})\ge\bar\alpha_i^N$.
\end{lemma}
\begin{proof}
By~\Cref{assum:schedule}, we have
$\overline Q_i=\bar\alpha_iI+(1-\bar\alpha_i)\frac1K\mathbf 1_K\mathbf 1_K^\top$. Hence, under
$\phi_i(\cdot)$~\eqref{eq:proposal}, each row independently selects the same
category $\xbf^{(i)}_r$ with probability $\bar\alpha_i+\frac{1-\bar\alpha_i}{K}\geq \bar{\alpha}_i$, which proves (B). It selects each
other category with probability $\frac{1-\bar\alpha_i}{K}$, which proves (A).
\end{proof}

\begin{lemma}\label{lem:retention} Let~\Cref{assum:schedule} hold, and let $\xbf^{(i)}\in \Xcal_{N,K}$ be the solution at diffusion step $1\le i\le M$.  For any set of
samples $\Xcal_{S_1}$ and temperature $\lambda>0$, $\hat p_i$ from~\eqref{eq:new_combined} satisfies
\eqn{
\hat p_i(\xbf^{(i)}|\xbf^{(i)})\ge1-\frac{N\beta_i}{1-\bar\alpha_i}.
}
\end{lemma}
\begin{proof}
Fix $\xbf_s\in\Xcal_{S_1}$ and a row $r$, and let $k$ and $k'$ be defined such that $[\xbf^{(i)}]_{rk}=1$ and $[\xbf_s]_{rk'}=1$.
By~\eqref{eq:posterior}, under $q(\cdot\mid\xbf^{(i)},\xbf_s)$, row $r$ takes a category
$j\neq k$ with a probability
\eqn{
\sum_{j\neq k}\frac{[\overline Q_{i-1}]_{k'j}[Q_i]_{jk}}{[\overline Q_i]_{k'k}}
\le\frac{\beta_i/K}{(1-\bar\alpha_i)/K}=\frac{\beta_i}{1-\bar\alpha_i},
}
since, by~\Cref{assum:schedule}, $[Q_i]_{jk}=\beta_i/K$ for $j\neq k$,
$\sum_{j\neq k}[\overline Q_{i-1}]_{k'j}\leq 1$, and $[\overline Q_i]_{k'k}\ge(1-\bar\alpha_i)/K$. Since the rows of $q(\cdot\mid\xbf^{(i)},\xbf_s)$ are independent by~\eqref{eq:posterior} and $\beta_i/(1-\bar\alpha_i)\leq 1$,
\eqn{
q(\xbf^{(i)}\mid\xbf^{(i)},\xbf_s)\ge\Big(1-\frac{\beta_i}{1-\bar\alpha_i}\Big)^N
\ge1-\frac{N\beta_i}{1-\bar\alpha_i}
}
by Bernoulli's inequality. This holds for every $\xbf_s$, and $\hat p_i$ is a convex
combination of these components, so the bound also holds for $\hat p_i$.
\end{proof}

Now we provide the proof for~\Cref{thm:convergence}:

\begin{proof}
Let $p_i^\star:=\Pr(\xbf^{(i)}\in\Xcal^\star)$, and let $\Fcal_i$ be the $\sigma$-algebra generated
by $\xbf^{(M)}$ and all samples drawn at steps $M,\ldots,i+1$. Then, $\xbf^{(i)}$ is
$\Fcal_i$-measurable. Since every element of $\Xcal^\star$ attains the maximum of $p_0$ over $\Xcal_{N,K}$ and no
other element does, the selection step of~\Cref{alg:discrete_mbd} (line 7) gives
$\mathbf{x}^{(i-1)} \in \Xcal^\star$ if and only if
$\Xcal^\star \cap (\Xcal_{S_1}\cup\Xcal_{S_2}) \neq \emptyset$. We now proceed in three steps.

\textbf{(Step 1)} Let $h_i:=((1-\bar\alpha_i)/K)^N$. We have two cases. First consider $\{\xbf^{(i)}\notin \Xcal^\star\}\in \Fcal_i$.  
Given $\Fcal_i$, the samples in $\Xcal_{S_1}$ are i.i.d. from the proposal $\phi_i$~\eqref{eq:proposal}. Thus, for any optimal solution $\xbf^\star \in \Xcal^\star$,
by~\Cref{lem:proposal}(A), we have
\eqn{\label{eq:find_step}
\Pr(\xbf^{(i-1)}\notin \Xcal^\star\mid\Fcal_i)\le\big(1-\phi_i(\xbf^\star)\big)^{S_1}\le(1-h_i)^{S_1}=:b_i.
}
On the event $\{\xbf^{(i)}\in\Xcal^\star\}\in\Fcal_i$, \Cref{lem:proposal}(B) gives
$\phi_i(\xbf^{(i)})\ge\bar\alpha_i^N$, and given $(\Fcal_i,\Xcal_{S_1})$, the samples in
$\Xcal_{S_2}$ are i.i.d.\ from $\hat p_i(\cdot \mid \mathbf{x}^{(i)})$ with
$1-\hat p_i(\Xcal^\star \mid \mathbf{x}^{(i)})
 \leq 1-\hat p_i(\mathbf{x}^{(i)} \mid \mathbf{x}^{(i)})
 \leq N\beta_i/(1-\bar\alpha_i)$
by~\Cref{lem:retention}. Note that $$1-\bar\alpha_i^N=(1-\bar\alpha_i)\big(1+\bar\alpha_i+\bar\alpha_i^2+\cdots+\bar\alpha_i^{N-1}\big)\leq N(1-\bar {\alpha}_i).$$

Combining the above with (a) $S_1\ge S_2$ and (b)
$\beta_i=(\bar\alpha_{i-1}-\bar\alpha_i)/\bar\alpha_{i-1}\le L/(Ma_{\min})$ by~\Cref{assum:schedule}, we get
\seqn[\label{eq:keep_step}]{
\Pr(\xbf^{(i-1)}\notin\Xcal^\star\mid\Fcal_i)
&\le (1-\bar\alpha_i^N)^{S_1}\Big(\frac{N\beta_i}{1-\bar\alpha_i}\Big)^{S_2} \\
&\le (N^2\beta_i)^{S_2}\le (\rho/M)^{S_2}.
}

With~\eqref{eq:find_step} and~\eqref{eq:keep_step}, we get
\seqn[\label{eq:combining_probs}]{
\Pr(\xbf^{(i-1)} \in\Xcal^\star\mid\Fcal_i)&\geq (1-b_i)\mathbb I\{\xbf^{(i)}\notin\Xcal^\star\} + \\
& \ (1-\ell)\mathbb I\{\xbf^{(i)}\in\Xcal^\star\},}
where $1-\ell=[1-(\rho/M)^{S_2}]_+$. Taking expectations in~\eqref{eq:combining_probs},
\eqn{\label{eq:recursion}
p_{i-1}^\star\ge(1-b_i)(1-p_i^\star)+(1-\ell)p_i^\star.
}

\textbf{(Step 2)} Since $0\le1-\ell\le1$, we can get
\seqn[\label{eq:recursion_monotone}]{
p_{i-1}^\star&\ge(1-\ell)(1-b_i)(1-p_i^\star)+(1-\ell)p_i^\star \\
&=(1-\ell)\big(1-b_i(1-p_i^\star)\big).
}

Since the right-hand side of~\eqref{eq:recursion_monotone} is nondecreasing in $p_i^\star$, with a
lower bound on $p_i^\star$, we can also lower bound $p_{i-1}^\star$.

Let $B_i:=\prod_{j=i+1}^{M}b_j$,
with $B_M:=1$. We show $p_i^\star\ge(1-B_i)(1-\ell)^{M-i}$ by induction on $i=M,M-1,\ldots,0$.
The case $i=M$ holds since $p_M^\star\ge0$. If it holds for $i$, then
by~\eqref{eq:recursion_monotone},
\seqn{
p_{i-1}^\star&\ge(1-\ell)\big(1-b_i+b_i(1-B_i)(1-\ell)^{M-i}\big)\\
&\geq(1-\ell)(1-b_iB_i)(1-\ell)^{M-i} \\
&=(1-B_{i-1})(1-\ell)^{M-i+1},
}
and thus the induction holds on $i=M,M-1,\dots, 0$. 

\textbf{(Step 3)} Using the result from Step 2, with $\Ical=\{1,\dots, M\}$, we get 
\eqn{\label{eq:product_bound}
\Pr\big(\xbf^{(0)}\in\Xcal^\star\big)=p_0^\star\ge\Big(1-\prod_{i\in\Ical}b_i\Big)(1-\ell)^M.
}

Finally, we bound $(1-\prod_{i\in\Ical}b_i)$ in~\eqref{eq:product_bound}:
\eqnN{
\prod_{i\in\Ical}b_i&=\prod_{i\in \Ical} \exp\left(S_1 \ln(1-h_i)\right) \\ 
&=
\exp\Big(S_1\sum_{i\in\Ical}\ln(1-h_i)\Big).}
Since $\ln(1-h)$ is concave and decreasing in $h$, Jensen's inequality gives
\eqnN{
\prod_{i\in\Ical}b_i&=\exp\Big(S_1M\cdot\frac1M\sum_{i\in\Ical}\ln(1-h_i)\Big)\\
&\le\exp\Big(S_1M\ln\Big(1-\frac1M\sum_{i\in\Ical}h_i\Big)\Big).
}

Since $(1-a(i/M))^N$ is nondecreasing with respect to $i \in \Ical$ by~\Cref{assum:schedule}, we get
$(1-a(i/M))^N\ge M\int_{(i-1)/M}^{i/M}(1-a(\tau))^Nd\tau$. Then using
the additivity of the integral, we know $\sum_{i\in \Ical} h_i\geq M \gamma_N/K^N$. Therefore, 
\eqn{\label{eq:final}
\prod_{i\in\Ical}b_i\leq \Big(1-\frac1M\sum_{i\in\Ical}h_i\Big)^{S_1M}
\le\Big(1-\frac{\gamma_N}{K^N}\Big)^{S_1M}.
}
By Bernoulli's inequality, $(1-\ell)^M\ge1-M\ell$ since $0\leq \ell\leq 1$ and $M\geq 1$. Together with $(1-\ell)^M\ge0$ and $\ell\le(\rho/M)^{S_2}$, this gives $(1-\ell)^M\ge\big[1-\rho(\rho/M)^{S_2-1}\big]_+$.
Substituting this and~\eqref{eq:final} into~\eqref{eq:product_bound} completes the proof.
\end{proof}

\bibliographystyle{plainnat}
\bibliography{references}

\end{document}